# CNN-based local features for navigation near an asteroid

**Olli Knuuttila[1], Antti Kestilä[2], and Esa Kallio[1]**

[1] Aalto University School of Electrical Engineering, PL 11000, 00076 Aalto, Finland
[2] Finnish Meteorological Institute, PL 503, 00101 Helsinki, Finland

Corresponding author: Olli Knuuttila (e-mail: olli.knuuttila@aalto.fi).

**ABSTRACT** This article addresses the challenge of vision-based proximity navigation in asteroid exploration missions and on-orbit servicing. Traditional feature extraction methods struggle with the significant appearance variations of asteroids due to limited scattered light. To overcome this, we propose a lightweight feature extractor specifically tailored for asteroid proximity navigation, designed to be robust to illumination changes and affine transformations. We compare and evaluate state-of-the-art feature extraction networks and three lightweight network architectures in the asteroid context. Our proposed feature extractors and their evaluation leverage synthetic images and real-world data from missions such as NEAR Shoemaker, Hayabusa, Rosetta, and OSIRIS-REx. Our contributions include a trained feature extractor, incremental improvements over existing methods, and a pipeline for training domain-specific feature extractors. Experimental results demonstrate the effectiveness of our approach in achieving accurate navigation and localization. This work aims to advance the field of asteroid navigation and provides insights for future research in this domain.

## I. INTRODUCTION

ASTEROID exploration missions, such as the Hayabusa-2 [1] and OSIRIS-REx missions [2], have demonstrated the significance of vision-based proximity navigation in complex and dynamic environments. The emerging industry of on-orbit servicing necessitates proximity navigation, which shares many aspects with navigation in close proximity to asteroids.

The upcoming Hera [3] mission to the asteroid 65803 Didymos and its accompanying cubesats (Milani and Juventas) has been of particular interest to the authors due to our involvement in the Milani precursor cubesat APEX [4]. Although the APEX project was discontinued due to changes in the programme, the insights gained from it remain valuable.

Vision-based navigation near asteroids or satellites presents a unique challenge due to the limited amount of scattered light that illuminates the object, resulting in significant appearance variations depending on the direction of sunlight. Traditional feature extractors such as Scale-Invariant Feature Transform (SIFT) and Oriented FAST and Rotated BRIEF (ORB) cope poorly with such drastic changes in appearance. Addressing this challenge requires the development of a robust and efficient feature extraction method to enable accurate navigation and localization in space.

Feature extraction plays a crucial role in most vision-based navigation methods in the field of robotics, including simultaneous localization and mapping (SLAM), as well as absolute navigation approaches like on-board rendering-based Synthetic Photometric Landmarks (SPLs) [5]. Additionally, SLAM methods can incorporate pre-built feature databases. Relative navigation methods also benefit from feature extraction, particularly in environments where rapid changes occur compared to the amount of relative movement present. For instance, when a spacecraft orbits an asteroid, the asteroid's rotation can be significantly faster than the spacecraft's orbital velocity. Visual odometry techniques that follow the rotating asteroid generate a relative path much longer than the spacecraft's actual travel along its orbit, resulting in poor accuracy even over short orbital paths. By matching features across asteroid rotations, the error per distance traveled can be substantially reduced.

Another scenario where feature extraction is crucial is when orbiting the L4/L5 points of a binary asteroid system such as Didymos. The spacecraft's location relative to the secondary body changes much slower than the secondary body's appearance due to its orbit around the primary body.

In such cases, visual odometry based solely on optical flow would not be effective.

In this work, our primary objective is to develop a lightweight feature extractor specifically tailored for asteroid proximity navigation. We aim to address the challenges posed by asteroid environments by designing a feature extraction algorithm that exhibits invariance to illumination changes, moderate rotations, scaling, and affine transformations. Furthermore, we aim to compare and evaluate different local feature algorithms based on their mean matching accuracy (MMA), ratio of correct matches to ground truth matches (M-Score), spatial accuracy of correct matches, and orientation error when the matches are used to estimate relative pose, assuming that the feature 3D coordinates are known.

The main contributions of this work are as follows:

- A trained lightweight feature extractor specifically designed for asteroid proximity navigation.
- Improvements over high-performance state-of-the-art feature extractors, particularly in the context of asteroid navigation.
- Comparison of two state-of-the-art feature extraction algorithms and three lightweight network architectures in the asteroid context.
- A pipeline for training feature extractors specialized in a given domain.
- A training dataset derived from images taken during four missions to different small Solar System bodies.

Our code together with the trained models are available at https://github.com/oknuutti/navex, while the data used [6] is published through Zenodo.

The rest of this paper is organized as follows: Section II provides an overview of related work. Section III presents the methodology and experimental setup, including data augmentation, evaluation metrics and hyperparameter optimization. Section IV provides details about image data preprocessing and the resulting datasets. Section V presents the results and performance analysis of our proposed feature extractor. Finally, Section VI concludes the paper and discusses potential future research directions.

## II. RELATED WORK

### A. FEATURES FOR NAVIGATION NEAR AN ASTEROID

Before discussing existing methods for local feature extraction that detect salient features and describe them using vectors of a certain length (descriptor/embedding), we will briefly review somewhat similar methods proposed for proximity navigation in space.

Traditionally, terrain relative navigation (TNR) near asteroids involved creating textured 3D maplets of small salient regions (natural landmarks) on the asteroid. This approach utilized a priori knowledge of the expected relative pose and direction of light to render the maplets, followed by template matching to locate them in the query image [7]. For the Rosetta mission, all the processing was performed on the ground by mission operators [8]. However, during the OSIRIS-REx mission, maplet rendering and template matching could be done either on the ground or automatically on board, but maplet creation was always performed on the ground [9]. An alternative method employed by Hayabusa-2 was the deployment of bright balls called target markers on the asteroid, which could subsequently be used as features [10].

Due to the challenges of creating maplets on board and the limited performance of traditional photometric features, different approaches have been suggested. One such approach is Synthetic Photometric Landmarks (SPLs) [5], which involves rendering a global shape model using a priori information. Traditional AKAZE features are then extracted and matched between the query image and the synthetic image. The AKAZE features perform adequately due to matching lighting conditions. However, creating the global shape model on board remains a non-trivial task and would likely be performed on the ground instead.

Convolutional neural networks (CNNs) have been extensively used in studies to automatically detect and describe craters for use as landmarks in navigation [11]–[14]. However, recent visits to small solar system bodies suggest that sub-kilometer objects do not possess many suitable craters for this purpose.

For on-orbit rendezvous with an uncooperative spacecraft, a common approach is to encode a few tens of target object features in the network weights. This allows a CNN classifier to output a heat map for each trained feature [15]–[18]. This approach provides the benefit of estimating feature location uncertainty, which can be subsequently used by the navigation filter. However, this approach requires prior knowledge of the target shape model, and the computational performance of the network degrades with each additional feature included.

In the context of asteroid navigation, there are methods that use deep learning to regress the center of volume, sub-solar point, and various points on the limb. This enables the extraction of pseudo range for subsequent use by the navigation filter [19], [20].

Pugliatti and Topputo [21] utilize a CNN classifier trained on the target asteroid to determine the approximate position of the spacecraft. They then refine the relative pose solution using custom template matching. Instead of directly using navigation camera images as input, they employ a custom U-Net-type architecture derived from MobileNetV2 to preprocess the images into segmentation maps. Each pixel is classified to belong to boulders, craters, the background, the terminator, or the rest of the asteroid surface. This method assumes nadir-pointing images taken within a specific distance range and oriented so that the asteroid rotation axis points upwards in the image frame.

Mancini *et al.* [22] employ a network that calculates a per-image (global) descriptor from a central patch of a nadir-pointing image. The extracted descriptor is compared using L2-distance to a reference map of precomputed global descriptors spanning the area of interest on the target object.

A heat map is then generated, incorporating the results from odometry to indicate the most likely spacecraft location. Similar assumptions as those made by Pugliatti and Topputo [21] apply to the images used for navigation. However, it is not necessary to train the network specifically on the target body.

## B. CNN-BASED LOCAL FEATURES

The field of local feature detection and description has a rich history and encompasses both traditional and deep learning-based methods. Comprehensive surveys by Csurka *et al.* [23] and Chen *et al.* [24] cover the topic, with the latter focusing on deep learning for localization and mapping, including feature extraction methods. Jin *et al.* [25] introduce a benchmarking framework to facilitate comparisons of local feature extraction methods for relative pose estimation of wide baseline image pairs.

Various approaches have been developed for local feature extraction. Some methods utilize small patches around keypoints detected by external feature detectors, such as the traditional difference of Gaussians (DoG). Examples include HardNet [26], SOSNet [27], GeoDesc [28], and ContextDesc [29]. Key.net [30] focuses exclusively on feature detection by combining gradient-based detection with subsequent CNN layers. On the other hand, methods such as D2-Net [31], ASLFeat [32], D2D [33], and UR2KiD [34] compute dense descriptors and use processing steps in the descriptor space to detect a sparse set of features, sometimes leveraging middle CNN layers. SuperPoint [35], HF-Net [36], R2D2 [37], and DISK [38] directly compute both dense descriptors and detection scores.

To the best of our knowledge, there are only two previously published works that employ learning-based feature extractors for navigation in the proximity of small solar system bodies. Beccari [39] trains a SuperPoint feature extractor using the related MagicPoint teacher network [35], using images from the MS-COCO dataset and synthetic images of Eros and Bennu. The author compares the resulting SuperPoint extractors with traditional methods like SIFT, concluding that the latter outperform the former.

In contrast, a concurrent study by Driver *et al.* [40], published during the final stages of our research, utilizes ASLFeat as the base feature extractor trained with real data acquired from 16 small bodies across eight separate missions. This study reports superior performance of the proposed feature extractor compared to traditional methods. In the Conclusion section of our paper (Section VI), we will briefly discuss their results in relation to our findings.

Considering the available methods and the benchmark by Jin *et al.* [25], our focus lies on SuperPoint, HF-Net, R2D2, and DISK as potential candidates for our specific use case. Table 1 provides key details of these methods.

SuperPoint [35] adopts a VGG-style architecture with three 2x2 non-overlapping max-pooling layers to reduce computation. The resulting feature map has cells of size 8x8 pixels. The model comprises two heads: one for descriptors and one for feature detection. The detection head consists of a hidden 3x3 256-channel convolution layer followed by a 1x1 convolution layer with 65 channels, generating a full-sized map of feature salience through a soft-max operation and reordering. The descriptor head has also a hidden 3x3 256-channel convolution layer followed by a 1x1 convolution layer with 256 channels producing L2-normalized descriptors that are up-sampled to the original image size. SuperPoint is trained using a teacher model called MagicPoint, which has been trained on synthetically warped data with ground truth pixel correspondences, employing a loss function combining cross-entropy and hinge losses for feature detection and descriptor matching.

HF-net [36], a lightweight variant of SuperPoint, employs the first seven layers of the MobileNetV2 architecture [44] as its backbone. This variant reduces the hidden layer channels in the detector head and utilizes the remaining part of MobileNetV2 to calculate a global image descriptor for building and querying a global index, which can be used for relocalization/loop-closing in a SLAM system such as ORB-SLAM2 [45] or VINS [46]. HF-net is trained with teacher models for all outputs: SuperPoint serves as the teacher for the detector and local descriptors, while NetVLAD [47] serves as the teacher for the global descriptor. The training scheme employs multi-task learning [48] with adjusted weights for different loss terms.

R2D2 [37] jointly trains a local feature detector and descriptor. Its feature detection is divided into two aspects: repeatability, which ensures consistent detection across differently warped images, and reliability, which measures the likelihood of correct matches. The R2D2 architecture resembles VGG but replaces the max-pooling layers with dilated convolutions to maintain feature map resolution at the expense of increased computational load. The resulting feature map serves as the descriptor output. The repeatability and reliability heads process the squared feature map using 2-channel, 1x1 convolution layers followed by soft-max operations. The R2D2 loss function comprises three equally weighted terms: two from the repeatability head promoting local similarity and peakiness, and one based on a differentiable approximation of average precision (AP) for descriptor ranking.

The DISK architecture [38] is a variation of the original U-Net [49]. It uses a single convolutional layer per block instead of two, instance- instead of batch-normalization, and PReLU instead of ReLU non-linearities. The last 129-channel layer of the U-Net is split to produce the descriptor and the detector outputs. The descriptor part is L2-normalized, while the detector part is left unchanged. The cost function takes on a reinforcement learning perspective, where the network implements a probabilistic policy that is trained to maximize a simple reward function, which rewards correct matches and penalizes incorrect ones.

## III. PROPOSED APPROACH

We were particularly interested in the HF-net architecture for our use case and intended to closely follow its design and learning scheme. However, HF-net relies on SuperPoint as the teacher network for local features, which is trained on the MS-COCO dataset and NetVLAD for global descriptors, trained on the Google Street View Time Machine. Since no existing feature extraction networks are trained on asteroid imagery, our first task was to train our own teacher network. While HF-net utilizes MobileNetV2 as its backbone, we also considered MobileNetV3 [50] and EfficientNet [51] for our proposed feature extractor named the Light-weight Asteroid Feature Extractor (LAFE).

**TABLE 1.** Prominent CNN-based methods for dense feature extraction

| Method | Year | Input | Desc dim | Architecture | Training scheme | Training datasets |
| --- | --- | --- | --- | --- | --- | --- |
| SuperPoint | 2018 | Mono | 256 | VGG-style backbone, heads have a hidden convolution layer and a convolution output layer, reshape detection result, interpolate descriptor result | Detector: bootstrapping with a simpler detector. Descriptor: negative and positive pairs, hinge loss | MS-COCO [41] |
| HF-Net | 2019 | Mono | 256 | Same as SuperPoint but change backbone to first layers of MobileNetV2, only 128-channel hidden layer for detector head | Same as SuperPoint but use SuperPoint as the teacher model | Aachen Day-Night [42], RobotCar Seasons [42], CMU Seasons [42] |
| R2D2 | 2019 | RGB | 128 | VGG-style backbone but without max-pooling, instead dilation is used at certain convolution layers thus preserving input size across the net | Detector: peakiness, cosim. Descriptor: mean AP loss | Processed Aachen and web images [37] |
| DISK | 2020 | RGB | 128 | U-Net with four down- and up-blocks, each with a single 5x5 convolutional layer, the last layer has 129 channels separated into descriptor and detector outputs | Reinforcement learning-derived loss function with rewards/penalties given for true/false matches | A subset of the MegaDepth dataset [43] |

For the teacher network, which we will refer to as the High-performance Asteroid Feature Extractor (HAFE), we evaluated both R2D2 and DISK. SuperPoint was excluded due to its suboptimal training scheme requiring a teacher network. After preliminary testing, we found that the U-Net backbone of DISK outperformed the VGG-style backbone of the R2D2 network. Therefore, we also adopted a U-Net backbone for the R2D2 network, which we refer to as R2D2-U.

It is important to consider that the majority of CNN architectures do not possess inherent scale or rotation invariance. Although some resilience against scale changes and rotations can be achieved through appropriate data augmentation, this form of invariance is acquired through brute force, consequently consuming network capacity. A more effective approach to achieve scale invariance involves extracting features at various scales during test time. This is accomplished by constructing an image pyramid with a specific scaling interval and then feeding each scale into the feature extraction network. Traditional methods like SIFT and ORB also adopt this approach.

In certain applications, rotation invariance might not be a prerequisite, especially when input images consistently exhibit a particular orientation. For instance, images from self-driving cars are typically captured horizontally, keeping the sky consistently at the top. Drones, when equipped with downward-facing cameras, can potentially leverage magnetometer readings to align images with north as the upper direction. Similarly, asteroids with stable rotation axes can adopt a similar strategy by utilizing star-tracker readings. Alternatively, images can be rotated based on the position of the Sun. If these input-conditioning techniques prove inadequate, achieving rotation invariance can be pursued through a spatial transformer [52]. However, for this study, we presumed that rotation-axis-based input conditioning would suffice, leaving the exploration of a spatial transformer's integration for future investigation.

In order to narrow the scope of this study, our focus was exclusively on local feature extraction, omitting components necessary for global feature extraction. Following the approach of HF-net, we employed grayscale images as input, generating 128-dimensional floating-point descriptors as part of the output. The assessment of RGB image inputs presents challenges, given the prevalent use of grayscale imagery in available asteroid data. The evaluation of binary descriptors was intentionally deferred to a subsequent phase to further constrain the study's scope.

The evaluation of various design choices for both LAFE and HAFE was difficult as the outcome is affected by a large number of hyperparameters. These hyperparameters are free parameters related to data augmentation, cost function, training, and the network itself. For instance, the effectiveness of a specific backbone architecture may hinge on an appropriate weight decay value. To tackle this challenge, we optimized these parameters using the Ray Tune framework [53], which integrates the ASHA scheduling algorithm [54] and Bayesian Optimization [55]. Bayesian Optimization, in this context, utilizes Gaussian Processes to forecast network performance based on hyperparameter values, facilitating suggestions for new trial configurations. Ray Tune also supports parallel execution of trials across multiple machines and incorporates error recovery mechanisms. The ASHA scheduler aids in the early termination of trials that do not demonstrate top-tier performance, while accommodating concurrent trials. However, it is important to note that ASHA may favor parameter values that lead to good initial performance rather than optimal performance after full training. Consequently, certain parameters such as learning rates were excluded from the optimization process and were determined through less rigorous methods. Moreover, hyperparameters that exhibited negligible impact during preliminary testing were omitted from the optimization to limit the dimensionality of the search space.

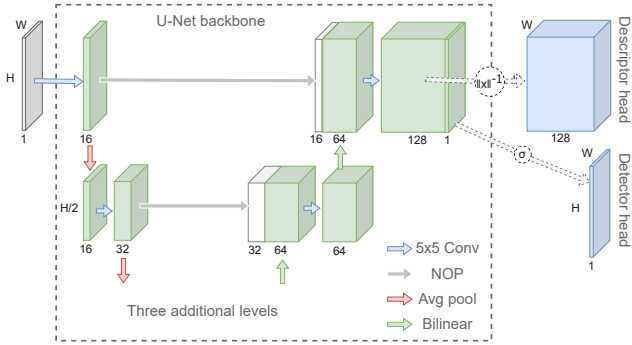

**FIGURE 1.** DISK architecture. The U-Net backbone is simplified in the figure by omitting the three deepest levels. See the text for additional details.

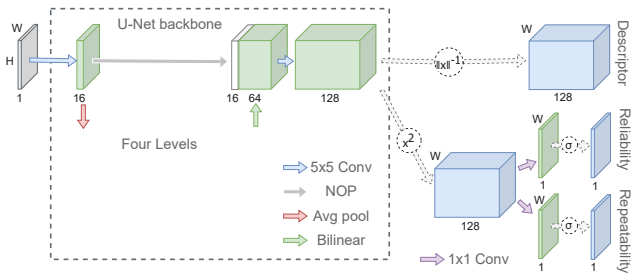

**FIGURE 2.** R2D2 with the same U-Net backbone (R2D2-U) than in DISK. Note that the figure shows one less level of U-Net than Fig. 1

### A. HIGH-PERFORMANCE ASTEROID FEATURE EXTRACTOR (HAFE)

#### 1) Architecture

From the outset, it was unclear which feature extractor was better, R2D2 with a U-net backbone (R2D2-U) or DISK. Consequently, an evaluation of both was deemed necessary. The principal distinction between R2D2 and DISK, when utilizing the same backbone, lies in their respective loss functions and output heads. Unlike DISK, which employs a descriptor and detection head, the R2D2 loss function distributes detection into two heads referred to as "repeatability" and "reliability." Within R2D2, both detection heads are fed the squared output of the descriptor head (as illustrated in Fig. 2), whereas DISK directly derives descriptors and detection scores from the final layer of the backbone (as depicted in Fig. 1).

The U-net backbone employed is identical to the original one used by DISK, comprising four down-blocks and up-blocks, each with a single 5x5 convolution layer, PReLU nonlinearities, and instance normalization. Down-sampling employs 2x2 average pooling, while up-sampling uses bilinear interpolation. Preliminary testing demonstrated inferior performance upon reducing down and up block count to three or utilizing two convolution layers per block. Conversely, elevating the count to five yielded negligible performance gains. The last layer comprises 129 channels (128 descriptor channels + 1 detector channel) when employed by DISK. However, in the context of R2D2-U, the last layer has only 128 channels.

The R2D2-U heads consist of a single 1x1 convolution layer featuring L2-normalization for descriptors and a special non-linearity for the detection heads:

$$f(x) = \frac{\log(1 + \exp(x))}{\log(1 + \exp(x)) + 1}. \quad (1)$$

In preliminary testing, we explored both R2D2 with output directly from the final layer of the backbone and DISK with R2D2-style head configuration. Both altered feature extraction approaches exhibited inferior performance compared to their corresponding baselines, prompting the retention of their original head structures. Nevertheless, we observed a performance enhancement upon reducing the channel count of R2D2's reliability head from two to one and employing the same single-channel activation function as the repeatability head [as defined in (1)] instead of the previous two-channel softmax.

We experimented with employing a sigmoid function, yet it appeared to degrade the performance of R2D2-U. Similarly, DISK's performance was improved by replacing the sigmoid detection score function with the function given by (1).

#### 2) Loss functions

The training process for both R2D2 and our modified version of DISK relies on utilizing image pairs with established pixel correspondences. In contrast, the original DISK training employed image triplets with pixel correspondences derived either from depth maps and camera poses, or through the application of epipolar constraints when depth maps were unavailable. Our training methodology involves batches of multiple image pairs, all of which are processed by the same network. We will begin by going through the R2D2 loss function, after which we continue with the DISK loss function.

*R2D2's loss function* aims to promote the sparse detection of highly distinctive and precisely localized features. The function comprises three key components: one that combines the descriptor and reliability outputs ($L_{AP\kappa}$), and two other components that are related to the repeatability output:

$$L_{R2D2} = L_{AP\kappa} + aL_{cosim} + bL_{peaky}, \quad (2)$$

where $a$ and $b$ are weights that could be optimized, in contrast to the original R2D2 publication by Revaud *et al.* [37], where they were statically set as $a = b = 1$. To facilitate optimization of these parameters, we find it more intuitive to reparameterize $a$ and $b$ as follows:

$$\begin{aligned} a &= 2(1-\beta)\alpha, \\ b &= 2\beta\alpha, \end{aligned} \quad (3)$$

where $\alpha > 0$ is the weight given to repeatability, while $\beta \in [0, 1]$ is the weight given to peakiness at the expense of cosine similarity. The corresponding values for the original R2D2 are then $\alpha = 1$ and $\beta = 0.5$.

In the original R2D2 formulation, all three loss terms exhibit variations within the range of 0 to 1 due to the addition of a constant value of 1 to each term. However, since constant terms bear no influence on the training process, we have omitted them. Consequently, the loss terms now fluctuate within the range of -1 to 0.

The loss term for cosine similarity, denoted as $L_{cosim}$, guides the repeatability output to exhibit local similarity across the image pairs. This term can be expressed as:

$$L_{cosim} = -\frac{1}{|P|} \sum_{p \in P} \frac{s_p \cdot s'_p}{\|s_p\| \, \|s'_p\|}, \quad (4)$$

where $P$ is a set of overlapping patches of size $n_{rep} \times n_{rep}$, extracted from the repeatability output of the first image within each pair, and subsequently flattened into vectors denoted as $s_p$. Conversely, vectors $s'_p$ are drawn from the second image's repeatability values, leveraging the known pixel correspondences. The patch size, dictated by $n_{rep}$, directly impacts the frequency of local maxima in the repeatability output. If certain pixels within $s_p$ lack corresponding matches, the corresponding repeatability values within $s'_p$ are set to those located at the bottom-right corner of the repeatability map. This approach, employed in the original R2D2 study, proves superior to entirely discarding these values from both patches.

The loss term for "peakiness," denoted as $L_{peaky}$, serves to enforce the sparsity of the repeatability output while discouraging the trivial solution of a constant value, which is permitted by the cosine similarity. This term can be expressed as:

$$L_{peaky} = -\frac{1}{|R|} \sum_{r \in R} [max(s_r) - mean(s_r)], \quad (5)$$

where $R$ is a set of patches $s_r$, extracted from a sliding window of size $n_{rep} \times n_{rep}$ from the repeatability output of all training images. Notably, the output of the first and second images within each pair is treated individually and equivalently. The *max* function returns the highest value within each patch, while the *mean* function computes the average value of a given patch.

Finally, the term aimed at optimizing Average Precision (AP) can be written as:

$$L_{AP\kappa} = -\frac{1}{|Q|} \sum_{q \in Q} [AP(q)R_q + \kappa(1 - R_q)], \quad (6)$$

where $Q$ represents a set of query descriptors sampled from the first image within each image pair. Here, $R_q$ denotes the reliability output of the descriptor $q$, $\kappa$ defines the threshold for acceptable AP, and $AP(q)$ stands for a differentiable approximation of the actual AP. The original formulation employed a fixed $\kappa = 0.5$. However, we observed that an initial warm-up phase (comprising 1500 steps) during which $\kappa$ gradually increases to its final value improves training performance.

$AP(q)$ provides an approximation of the precise AP, which is calculated by initially populating an array with cosine similarity scores between descriptor $q$ and the descriptors within a set $B$ sampled from the second images across all pairs in the training batch. Next, a label array is created, with cell values being 0 except for the cell corresponding to the correct match, which is assigned a value of 1. This label array is arranged in descending order based on the similarity score, followed by a computation of the cumulative sum. The average of this cumulative sum yields the AP. The approximation circumvents the non-differentiable sorting process by quantizing the descriptor distances into a specified number of bins, assigning values between 0 and 1 depending on the proximity of the distance value to the bin center. The precise mathematical formulation for the AP calculation can be found in Balntas e*et al.* [56], while the details of the approximation can be found in He *et al.* [57].

Due to memory limitations, a subset of descriptors is used for each image. Specifically, we randomly sample one query descriptor $q$ for every 64 available descriptors. The positive match in the corresponding second image is identified by determining the closest matching descriptor within a radius $r_{pos}$ from the ideal pixel, thus accommodating some degree of error in pixel correspondences. A set of challenging distractor descriptors is sampled within a circular region around the location of the ideal match, at a distance of $r_{neg}$ from the optimal positive match. Additional distractors are randomly sampled (also at a 1/64 ratio) across all the second images in the batch, with the exclusion of the circular region in the corresponding second image defined by $r_{neg}$.

*The loss function of DISK* [38] is derived from reinforcement learning, particularly the REINFORCE method [58], which seeks to maximize the expected reward $\mathbb{E}[R|\theta]$ given a policy parameterized by $\theta$. This entails stochastic gradient ascent, where the policy, denoted as a probability function $P(A|I, \theta)$, governs possible actions $A$ in relation to the input $I$ and policy parameters $\theta$. The fundamental approach involves iterative sampling of actions from the policy, evaluating the gradient at the sampled actions with respect to $\theta$, and then updating $\theta$ according to the gradient.

In DISK, the input $I$ is divided into image pairs ($I_A$ and $I_B$), each generating feature sets ($F_A$ and $F_B$, respectively). The set of possible actions $A$ is defined in a relaxed manner, allowing simultaneous matching of each potential feature pair between $F_A$ and $F_B$, meaning feature $i \in F_A$ can be concurrently matched with both $j, k \in F_B$ with non-zero probability. The policy probability function $P(i \leftrightarrow j|I_A, I_B, \theta)$ is factorized into a descriptor matching function and two identical feature detection functions:

$$P(i \leftrightarrow j | I_A, I_B, \boldsymbol{\theta}) = P(i \leftrightarrow j | \boldsymbol{\delta}_i, \boldsymbol{\delta}_j, \theta_M) \\ \cdot P(i|K_A) \cdot P(j|K_B), \quad (7) \\ (K_k, \boldsymbol{\delta}_k) = f(I_k, \boldsymbol{\theta}_w)_i, \quad k \in \{A, B\},$$

where $\boldsymbol{\delta}_i$ and $\boldsymbol{\delta}_j$ represent the descriptors of features $i \in F_A$ and $j \in F_B$, respectively, while $\theta_M$ indicates the scale of descriptor match L2-distances. Symbols $K_A$ and $K_B$ stand for feature detection maps of images $I_A$ and $I_B$, respectively. The function $f(I_k, \boldsymbol{\theta}_w)$ represents the DISK network.

Similar to R2D2, only a subset of features is sampled. In DISK, sampling entails: 1) dividing the detection output $K$ into cells $K^u$ of size $h \times h$ ($h = 8$); 2) employing *softmax* for probability normalization within each cell $K^u$; 3) randomly proposing one sample per cell based on normalized probabilities, and finally; 4) accepting each proposed sample $i$ with the probability given by the original detection output $K_i$. Consequently, the detection probability functions from (7) can be reformulated as:

$$P(i|K_k) = \text{softmax}(K_k^u)_i \cdot K_{k,i}, \quad k \in \{A, B\}. \quad (8)$$

The probability function for descriptor matching is factored into the forward matching $i \rightarrow j$ and backward matching $i \leftarrow j$ components, each normalized individually via *softmax*:

$$P(i \leftrightarrow j | \boldsymbol{\delta}_i, \boldsymbol{\delta}_j, \theta_M) = \text{softmax}(-\theta_M D_{i,\cdot})_i \\ \cdot \text{softmax}(-\theta_M D_{\cdot,j})_j, \quad (9)$$

where $D$ represents a distance matrix computed between each descriptor $\boldsymbol{\delta}_i$ and $\boldsymbol{\delta}_j$. The notation $D_{i,\cdot}$ extracts the $i$-th row from $D$, while $D_{\cdot,j}$ extracts the $j$-th column. The descriptor distance scale is given by $\theta_M$, which is the reciprocal of the softmax temperature.

With these foundational concepts established, we can delve into the overall loss function, expressed as:

$$L_{DISK} = L_{RE} + \lambda_{kp} L_{KP}, \quad (10)$$

where $L_{RE}$ stands for the loss component tied to the REINFORCE method, while $L_{KP}$ represents an additional feature detection cost serving as a regularizer. This cost is weighted by $\lambda_{kp}$ (the original DISK employs $\lambda_{kp} = 0.001$). The cost term is defined as:

$$L_{KP} = \sum_{i \in F_A} \log P(i|K_A) + \sum_{j \in F_B} \log P(j|K_B). \quad (11)$$

By reformulating the gradient estimator from [38], the REINFORCE loss component can be written as:

$$L_{RE} = -\sum_{i \in F_A} \sum_{j \in F_B} P(i \leftrightarrow j | \boldsymbol{\delta}_i^*, \boldsymbol{\delta}_j^*, \theta_M) R_{ij} \Gamma_{ij}, \\ \Gamma_{ij} = \log P(i \leftrightarrow j | I_A, I_B, \boldsymbol{\theta}), \quad (12)$$

where $R_{ij}$ denotes the reward associated with matching feature $i$ with feature $j$, while $\Gamma_{ij}$ corresponds to the logarithm of the function given in (7). For correct matches, $R_{ij} = \rho_{tp}$, for false matches, $R_{ij} = \rho_{fp}$, and for cases where pixel correspondence is missing for feature $i$, $R_{ij} = 0$. A feature match is considered correct if it lies within an $\epsilon$-pixel distance from the true corresponding pixel. In the original DISK formulation, the reward values were $\rho_{tp} = 1.0$ and $\rho_{fp} = -0.25$.

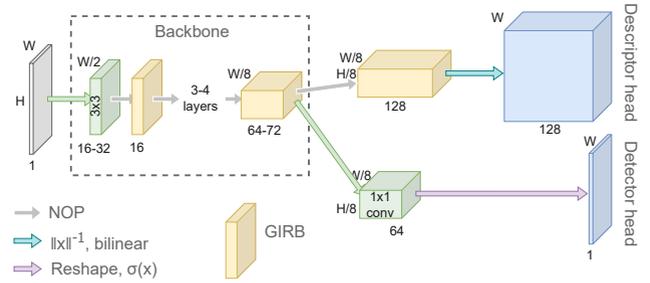

**FIGURE 3.** LAFE architecture. Various backbones that use GIRB as a building block (see Fig. 4) are considered.

Notably, $\boldsymbol{\delta}_i^*$ and $\boldsymbol{\delta}_j^*$ are detached copies of the descriptors, so that the gradient with respect to $\boldsymbol{\theta}_w$ is solely affected by the weight given by the probability function that the detached copies affect. In essence, this function assigns weight to matches with similar descriptors while largely diminishing the impact of matches with highly dissimilar descriptors. The detachment results in the anomaly of the loss potentially increasing during training, while simultaneously, the network's performance continues to improve as expected.

### B. LIGHT-WEIGHT ASTEROID FEATURE EXTRACTOR (LAFE)
#### 1) Architecture

The architecture of the lightweight feature extractor is independent of the choice of the teacher network. It consists of a single detection head. In case the R2D2-U network is selected as the teacher, the target output for the single detection head is computed as the product of the two R2D2 detection heads. The architecture, as depicted in Figure 3, follows the design of HF-net [36], which is itself based on SuperPoint [35]. In both networks, to mitigate the loss of spatial resolution in the backbone ($W/8 \times H/8$), the 65-channel detection head output is reorganized such that each cell covers an $8 \times 8$ region. Before this reorganization, the output undergoes a channel-wise softmax operation, followed by the removal of the extra "no-detection" channel. The descriptor head output then restores the full spatial resolution through bilinear interpolation.

In contrast to HF-net and SuperPoint, LAFE adopts the same activation function as R2D2, defined in (1), while also omitting the "no-detection" channel. Additionally, LAFE employs 128-channel descriptors, in alignment with DISK and R2D2.

We evaluated three distinct lightweight backbones: MobileNetV2 [44], MobileNetV3 [50], and EfficientNet-B0 [51]. These backbones were modified to support grayscale input and the channel count in their final layer was increased, a modification similar to that applied in HF-net.

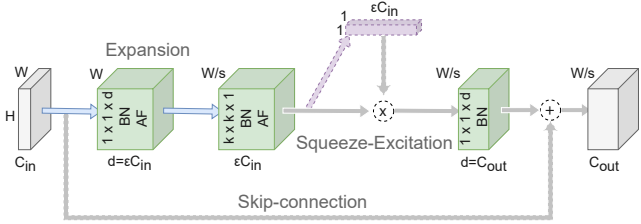

**FIGURE 4.** General inverted residual block (GIRB) used by different LAFE backbones as building blocks. The typical expansion factor $\varepsilon = 6$ and stride $s$ is 1 or 2. A skip-connection is used if output channel width $C_{out}$ equals input channel width $C_{in}$. Squeeze-excitation is optional. The activation function (AF) can be Hard-swish or ReLU6.

The HF-net backbone is based on MobileNetV2, with the channel counts of the last two layers increased from 32 to 64 and from 32 to 128, as per the source code referenced in the article [36]. However, the article itself cites the channel widths as 48 and 96. Since our descriptors only require 128 channels (instead of 256), we concluded that for MobileNetV2, elevating the last layer's channel count from 32 to 64 suffices. For both MobileNetV3 and EfficientNet-B0, we raised the channel count of the final layer from 40 to 72, approximating the geometric mean of 40 and 128.

Inspired by the block architectures within the various lightweight backbones, we propose replacing the first $3 \times 3$ convolution layer in the descriptor head with a generalized inverted residual block (GIRB), as illustrated in Figure 4. All three lightweight backbones can be implemented using this generalized block. For instance, MobileNetV2 omits Squeeze Excitation (SE) entirely, while MobileNetV3 employs it for some blocks, and EfficientNet uses it for all blocks. Furthermore, characteristics such as kernel size, stride, output channel count, expansion factor, and the chosen activation function vary.

During preliminary testing, we experimented with employing a GIRB for the detection head, but discovered that detection performed well without any hidden layers. Initially, we had planned to employ hyperparameter optimization to determine the specific details of the descriptor head's GIRB. However, it soon became evident that optimization consistently favored higher channel counts and expansion factors, even for marginal gains in performance, thus increasing the network's capacity and undermining the lightweight nature of LAFE. Ultimately, we limited hyperparameter optimization to choosing one of the three available backbones and deciding whether SE would be used in the descriptor head or not. The channel count and expansion factor for the descriptor head were fixed at 128 and 6, respectively.

2) Loss function

The loss function employed for LAFE closely follows that of HF-net, with the exception of excluding the global descriptor term:

$$L_{LAFE} = e^{-w_1} \sum_{i \in F} \|\boldsymbol{\delta}_i^s - \boldsymbol{\delta}_i^t\|_2^2 \\ + 2e^{-w_2} \sum_{i \in F} BCE\left(K_i^s, K_i^t\right) \quad (13) \\ + w_1 + w_2,$$

where $w_1$ and $w_2$ are weights associated with multitask learning [48], and they are jointly optimized with the network weights. The index $i \in F$ iterates through all the detection values $K^s$ and their associated descriptors $\boldsymbol{\delta}^s$. The matching target values are denoted as $K^t$ and $\boldsymbol{\delta}^t$. Unlike training the teacher network, no sampling is required.

In principle, LAFE could be trained using R2D2 or DISK loss functions. However, as argued by the authors of HF-net [36], learning to predict the output of a teacher network is a more straightforward learning task. This allows us to expect reasonable performance from LAFE, even though it is less capable than HAFE.

### C. DATA AUGMENTATION
An essential aspect of neural network training is data augmentation, where training data is transformed to retain essential information while altering non-essential aspects. This helps prevent overfitting and leads to improved performance on new data.

*Our data augmentation pipeline for HAFE* training data consists of operations that can affect either one or both images in any given image pair:

1) First image: Random scaling so that the shortest edge width is 256–1024 pixels.
2) Second image: The true relative scale between the image pair is inferred from pixel correspondences, and the second image is randomly scaled so that the scale difference $k_{rnd}$ between the pair is at most half of the image pyramid scaling factor $k$ used during inference, i.e., $k_{rnd} \in [k^{-1/2}, k^{1/2}]$.
3) First image: Random cropping weighted by available pixel correspondences in potential cropping areas.
4) Second image: Deterministic cropping that maximizes the number of pixel correspondences.
5) Both images: Random horizontal flip, either flipping both images or none.
6) Both images: Add uniformly distributed pixel noise with amplitude $\lambda_n$.
7) Second image: Random brightness change by multiplying the image with gain $g$ distributed as $\ln(g) \sim \mathcal{U}(-\ln(\lambda_g), \ln(\lambda_g))$, where $\mathcal{U}$ represents the uniform distribution.

We have chosen to follow R2D2 [37] and selected the image pyramid to have $s = 4$ images per octave (doubling of scale), making the scaling factor $k = 2^{1/s} \approx 1.189$. Depending on the final application, a trade-off analysis between resource usage and accuracy should be performed to select an optimal $s$.

For certain datasets lacking geometry backplanes that allow image pair construction, we generate image pairs by warping each image with a random homography before using the data augmentation pipeline outlined above. The homography transformation can be factored into the rotation, translation, shear, and projection components. Based on initial testing, we found that rotation and projection components seemed sufficient. The random transformation matrix used for the synthetic pairs becomes:

$$H_{rnd} = \begin{bmatrix} \cos(\phi) & -\sin(\phi) & 0 \\ \sin(\phi) & \cos(\phi) & 0 \\ 0 & 0 & 1 \end{bmatrix} \begin{bmatrix} 1 & 0 & 0 \\ 0 & 1 & 0 \\ p_1/w & p_2/h & 1 \end{bmatrix},$$
$$\phi^2 \sim \mathcal{U}\left(-\lambda_r^2, \lambda_r^2\right),$$
$$p_1^2, p_2^2 \sim \begin{cases} \mathcal{U}\left(0, \lambda_p^2\right), & \text{if } z = 1, z \sim B(0.5) \\ \mathcal{U}\left(\left[(\lambda_p + 1)^{-1} - 1\right]^2, 0\right), & \text{otherwise,} \end{cases}$$
(14)

where $B$ represents the Bernoulli distribution, $\lambda_r$, and $\lambda_p$ are hyperparameters determining the extremeness of the generated rotations and projections, and $w$ and $h$ are the image width and height in pixels, respectively. Note that the square of $\phi$, $p_1$, and $p_2$ are (piece-wise) uniformly distributed. This has the advantage of generating more extreme values, which tend to be more valuable for training purposes.

*LAFE training* does not require paired images, leading to a slightly different data augmentation pipeline:

1) Random rotation and projection $H_{rnd}$ (same as used for synthetic pairs).
2) Random scaling so that the shortest edge width is 256–1024 pixels.
3) Random cropping.
4) Random horizontal flip.
5) Uniform pixel noise.
6) Random exposure.

All these steps use the same hyperparameter values as the paired image pipeline. There is an additional opportunity to use data augmentation because a slightly modified image can be fed to the student network compared to the one given to the teacher network. Here, we consider using random exposure with a maximum gain $\lambda_g^{st}$, followed by adding normally distributed noise with a standard deviation $\lambda_\sigma^{st}$.

To reduce randomness in validation performance metrics, validation data is not processed by these pipelines. However, image scaling and cropping are still necessary. If the shortest edge width of the image is not in the 256–1024 pixel range, the image is either upscaled or downscaled so that the shortest edge becomes either 256 pixels or 1024 pixels. Central cropping is used for LAFE. For HAFE, both the first and second image cropping locations are chosen to maximize the pixel correspondence count of the resulting cropped image pair.

### D. METRICS

The matching pipeline employed for evaluating feature extractors proceeds as follows: it first extracts a sparse set of features from both images using non-maxima suppression (NMS), discards features with low detection scores, calculates a descriptor distance matrix between all retained descriptor pairs, and applies mutual nearest neighbor criteria to eliminate non-circular matches. Matches are labeled as "possible to match" if a pixel correspondence exists for the first image descriptor location. Additionally, if the matching second image descriptor is within 5 pixels of the ground-truth location, the match is labeled as correct.

In case a network has multiple detection outputs (as in R2D2), the repeatability output is employed for NMS. NMS is executed by initially filtering out the highest spatial frequencies using a $3 \times 3$ averaging kernel and then selecting all locations that are the maximum in their $3 \times 3$ neighborhoods. A feature is discarded after NMS if the corresponding detection output (values $\in [0, 1]$) falls below the threshold of 0.5 or if its combined detection score (repeatability times reliability) is lower than that of the top $N$ features, where $N = round(0.001hw)$, and $h$ and $w$ represent the height and width of the image, respectively.

During the training and hyperparameter optimization phases, features are extracted at a single scale level. However, when evaluating the final feature extractors, features are extracted at various scale levels by creating an image pyramid. Along with image coordinates, the features also retain the scale of the image from which they were extracted. To account for these feature scales during matching, we follow the methodology proposed in [59]. Matches are initially established without considering their scales. Subsequently, the final matches are determined by adjusting the scales based on the estimated intrinsic scale difference and constraining the matches to the nearest scale levels.

Several metrics can be derived from the matching pipeline to evaluate the performance of the feature extractors:

- The ratio of correct matches over all possible matches (Matching Score, *M-Score* [35], [37]), which serves as the target metric for hyperparameter optimization, as it is the only metric that cannot be improved simply by detecting fewer features.
- The ratio of correct matches over proposed matches (Mean Matching Accuracy, *MMA* [31], [37]), which quantifies the quality of proposed features without considering missed opportunities.
- Mean Average Precision (*mAP*, [56]), discussed previously in the context of the R2D2 loss function.
- Pixel Localization Error (*LE*, [35]), providing the average image-space distance between correct matches and their associated ground truth.

### E. HYPERPARAMETER SEARCH

The performance of the feature extractor designed in this study is highly dependent on the hyperparameter values inherent to the extractor network and its training process. Hyperparameter optimization is, therefore, indispensable if we hope to produce a state-of-the-art feature extractor. Various hyperparameter optimization frameworks are available, and in this study, we have chosen Ray Tune [53]. Ray Tune not only distributes the training workload across different computing nodes but also provides interfaces for exploring the search space and scheduling trials. A trial involves evaluating the search space at a specific point determined by the search method. The evaluation entails training the feature extractor for a specified number of epochs and calculating the target performance metric on the validation dataset. The trial scheduler manages which trials are processed by computing nodes and decides whether to pause or terminate a trial before it reaches the maximum number of epochs. For our search method, we have employed a Bayesian Optimization (BO) variant implemented by the Scikit-Optimize software package [55]. This approach is combined with the Asynchronous Successive Halving Algorithm (ASHA) scheduler [54], which enables us to prioritize promising trials while discontinuing non-promising ones early in the process. Both the Aalto University HPC cluster (Triton) and the CSC IT Center for Science HPC cluster (Puhti) were utilized during the study. Six computing nodes featuring NVIDIA Tesla V100 Volta GPUs were utilized in parallel during the optimization phase.

1) ASHA Scheduling

The core concept behind ASHA is to allocate a small initial resource budget, denoted as $r_0$ (e.g., one training epoch), to each trial. Subsequently, only the top-performing $1/\eta$ trials are allowed to continue with an increased budget of $r_1 = \eta r_0$ per trial. Trials are terminated when they reach the maximum resource usage per trial, denoted as $r_{max}$. Instead of measuring resource use in terms of epochs, we opted for 1500 training batches to obtain more frequent validation results and reach the first ASHA decision point sooner. With $\eta = 3$, $r_0$ set to 1500 training steps, and $r_{max}$ at 24000 training steps, stopping decisions are made at 1500, 4500, and 13500 steps. We maintained a fixed total of 243 trials, expecting at least 9 fully trained trials, although the asynchronous nature of the algorithm may yield a larger number. The approximate total resource usage can be estimated as $792r_0$.

The ASHA variant implemented by Tune differs slightly from the one described in [54]. In this variant, a decision is taken immediately whether to halt or continue a trial at each rung, as opposed to pausing trials for potential promotion later. This modification allows us to update the search algorithm with intermediate and lower-fidelity evaluation results from stopped trials. However, it's worth noting that Scikit-Optimize does not explicitly support multifidelity evaluations, as discussed by Klein *et al.* [60], something which is beyond the scope of our current work. In the same study, the authors also compared the original trial-promoting ASHA with the Tune variant featuring trial stopping, with the latter yielding superior performance [60].

2) Scikit-Optimize Bayesian Optimization

Scikit-Optimize's implementation of Bayesian Optimization (BO) compares favorably to other BO methods [61], even though some methods, such as Trust Region Bayesian Optimization (TuRBO) [62] and ensembles of TuRBO and Scikit-Optimize, outperform it to a certain extent. Our understanding of Scikit-Optimize is primarily derived from an analysis of its source code, as comprehensive articles on the topic were not readily available. Scikit-Optimize's BO recommends hyperparameter values for new trials by constructing a Gaussian Process (GP) surrogate model to predict network performance based on hyperparameter values. Whenever a new evaluation result becomes available, the prior surrogate model is discarded, and a fresh one is fitted using the expanded training data. The training process involves estimating GP kernel parameters, such as hyperparameter length scales and Gaussian noise, through maximum likelihood gradient ascent. These length scales offer valuable insights into the influence of specific hyperparameters on network performance; larger length scales correspond to lower impact. The parameter space is normalized to the 0–1 range, and the reported length scales are also presented in this normalized space.

After GP fitting, new suggestions for evaluating points in the search space are generated by randomly selecting one of three acquisition functions: Probability of Improvement (PI) [63], Expected Improvement (EI) [64], and Upper Confidence Bound (UCB) [65]. This strategy of combining multiple acquisition functions has been shown to be more effective than relying on any single criterion [66]. To mitigate the risk of converging to a local maximum, the surrogate model search involves sampling 10,000 random locations, with the best five serving as starting points for gradient ascent optimization. Generating multiple distinct points with identical knowledge is necessary due to parallel search space evaluation. This is achieved by assuming a poor dummy response for previously generated points that lack real responses [67].

Intuitively, Successive Halving, and by extension, ASHA, may exhibit a bias toward early performance at the expense of overall performance because only the early high performers receive full training. However, the extensive exploration of the search space made possible by ASHA outweighs this bias, as demonstrated by studies comparing random search ASHA to random search with full training [54], [68]. Moreover, although Scikit-Optimize treats intermediate, low-fidelity performance evaluations from trials stopped by ASHA as full evaluations, based on Wulff *et al.* [68], the combination of fixed-fidelity BO with ASHA outperforms random search ASHA. Consistent with Wulff *et al.* [68], we excluded the learning rate from optimization, given its potential to introduce early performance bias. To ensure that we did not overlook trials achieving their best performance early, we assessed performance based on the maximum M-Score reached by a trial during any validation run conducted after each 1500 training step interval.

3) Hyperparameter selection

All free parameters involved in the process of generating a feature extractor that are not optimized through backpropagation can be classified as hyperparameters. However, due to limited computational resources and the chosen optimization method, only a subset of possible hyperparameters can be optimized. To restrict the search space, most parameters related to network architecture, such as the number of layers, channel widths, activation functions, etc., have been excluded. Other optimization methods, which exploit weight sharing to reduce training times, may be better suited for network architecture search (NAS), but NAS was considered to be beyond the scope of this study. Additionally, we excluded parameters that are expected to have values within a reasonable range or minimal impact on the target performance metric.

The optimization outcome corresponds to the hyperparameter configuration that yielded the best performance. However, due to the stochastic nature of evaluations, we also extracted the best configuration as indicated by the surrogate model.

We focused initially on hyperparameter optimization of the two HAFE models, R2D2-U and DISK. The best-performing model among the two was subsequently chosen as the teacher for LAFE during its hyperparameter optimization. As the teacher network training did not include synthetic data, we also excluded it in the LAFE training. Section V Results presents the detailed hyperparameters optimized, their search spaces, and the optimization results. Before delving into the results, we will provide an overview of the image data used for training, validation, and testing.

## IV. ASTEROID/COMET DATA

Various image sets are available thanks to several exploration missions targeting solar-system small bodies. Notable recent endeavors include the OSIRIS-REx mission, which orbited and later sampled asteroid 101955 Bennu, and Hayabusa-2, which undertook a similar mission with asteroid 162173 Ryugu. Both missions successfully reached their target asteroids in 2018. The first mission that provided extensive imagery of an asteroid was NEAR Shoemaker, which orbited 433 Eros in 2000 and subsequently landed on its surface in 2001. Following this, the Hayabusa spacecraft imaged asteroid 25143 Itokawa while orbiting it in 2005. In 2014, Rosetta achieved the distinction of becoming the first spacecraft to orbit a comet, 67P/Churyumov–Gerasimenko.

The images acquired during these missions served as the basis for training our feature extractor. Additionally, we supplemented our dataset with synthetic data generated using a Bennu shape model [69] and OpenGL-based rendering software [5], [70]. Images from various missions can be accessed through the NASA Planetary Data System (PDS) or, in the case of the Rosetta mission, via the ESA Planetary Science Archive [71]. The datasets employed in this study are detailed in Table 2. It's worth noting that some missions encompass multiple instruments, each contributing to a distinct dataset.

Interpreting the available data requires caution, as each mission adopts its own image and metadata formats. None of the datasets provide readily available spacecraft-to-target-body relative pose information. To overcome this limitation, we estimated relative poses based on pixel georeferencing, camera instrument intrinsics, and employed them during image pair creation. For datasets such as 67P/Churyumov–Gerasimenko (67P/C–G) NAVCAM and Bennu TAGCAMS, which lack geometry backplanes, we opted for image warping to generate synthetic image pairs, as we did not pursue geometry estimation through structure-from-motion (SfM) algorithms. Images of Ryugu were not included in this study due to their late availability.

When dividing the datasets for training and validation, we excluded both the synthetic images and the synthetic pairs of real images from the validation set. This approach ensures that hyperparameter optimization focuses solely on real image pair performance.

The total count of available images, as presented in Table 2, only includes images that are georeferenced, if supported by the respective dataset. For datasets lacking georeferencing, the total count encompasses images taken during proximity operations, with those acquired during the cruise and approach phases being excluded. A selection process was applied to filter out images that are corrupted, saturated, or contain only a small portion of the target body. In cases where datasets contained a surplus of acceptable images, we randomly selected a subset to create a balanced combined dataset. Notably, Bennu TAGCAMS images often suffer from saturation due to the navigation mode's requirement to capture background stars. Due to the limited number of available Itokawa images, we reserved that dataset exclusively for testing purposes. Additionally, the datasets 67P/C–G OSIWAC and Bennu OCAMS were omitted due to time constraints imposed by our work schedule.

**TABLE 2.** Image Datasets

| Target body | Camera instrument | Ref. | Geo-ref. | Available | Acceptable | Selected | Pairs |
|---|---|---|---|---|---|---|---|
| Eros | MSI | [72] | yes | 93397 | 85960 | 8596 | 5836 |
| Itokawa | AMICA | [73] | yes | 766 | 766 | 766 | 1496 |
| 67P/C–G | NAVCAM | [74] | no | 16758 | 10238 | 5119 | synth |
|  | OSINAC | [75] | yes | 25461 | 19455 | 3891 | 6733 |
|  | OSIWAC | [75] | yes | 12346 | ? | - | - |
| Bennu | TAGCAMS | [76] | no | 24803 | 1698 | 1698 | synth |
|  | OCAMS | [77] | no | 27223 | ? | - | - |
| Synth. Bennu | - | - | yes | - | - | 15926 | 7963 |

### A. PREPROCESSING

The images utilized in this study encompass various processing levels, ranging from raw images to cleaned-up radiance factor (I/F) images that may have undergone resampling to correct for geometric distortions. As a compromise between efficient CNN training and image quality, we reduce the image depth to 8 bits and save them with lossless PNG compression. To mitigate information loss when encoding pixel values with only 8 bits, we calculate the percentiles of the image pixel values at $p_{lo} = 0.05\%$ and $p_{hi} = 99.99\%$, denoted as $v_{lo}$ and $v_{hi}$, respectively. We then rescale the pixel values based on these percentiles and apply gamma correction with $\gamma = 1.8$ as follows:

$$v' = 255 \left( \int_0^1 \frac{v - v_{lo}}{1.2 v_{hi} - v_{lo}} \right)^{1/\gamma}, \tag{15}$$

where $1.2 v_{hi}$ provides a margin to prevent highlights from saturating, and $\int_0^1$ indicates clipping to the 0–1 range. Images with a side length smaller than 256 pixels or those containing more than 1% completely black rows (e.g., some MSI images of Eros have these missing rows) are filtered out.

Following the rescaling procedure described in (15), we further filter out images that are ill-suited for feature extraction based on percentiles tuned to represent target size and image saturation level. For saturation level assessment, we employ $p_{lo}^{sat} = 99.8\%$ and $p_{hi}^{sat} = 99.99\%$. If $v_{hi}^{sat} - v_{lo}^{sat} \leq 5$, the image is considered saturated and is therefore discarded. Regarding target size estimation, we calculate the background level with the percentile $p_{bg}^{trg} = 4.0\%$. To determine a percentile corresponding to a minimum target size, we visualize a half circle with a radius $r$ and divide its area by the pixel count calculated from the image's width $w$ and height $h$:

$$p_{fg}^{trg} = 1 - \frac{\frac{1}{2}\pi r^2}{wh}. \tag{16}$$

A slightly different radius $r$ is employed for each dataset: $r = 75$ for Itokawa, $r = 100$ for Eros, $r = 115$ for 67P/C–G NAVCAM, $r = 185$ for Bennu, and $r = 230$ for 67P/C–G OSINAC. If the target appears too small in the image for our purposes, $v_{fg}^{trg}$ approaches the image background value. We assume this is the case if $v_{fg}^{trg} - v_{bg}^{trg} < 50$. Images where the target occupies the entire field of view typically exhibit sufficient shadows and contrast to pass the filtering process.

### B. PAIR FORMATION

Georeferenced images can be paired to generate realistic training data. Pairing is achieved by clustering the Cartesian coordinates of each georeferenced image pixel into four clusters using k-means clustering. The resulting cluster centroids are organized into a kd-tree [78], which can be queried to find nearby cluster centroids.

These centroid pairs are processed in random order. Images corresponding to centroid pairs are accepted as a pair based on criteria involving differences in viewing direction and distance. Centroid pairs referring to the same image or an already accepted pair are discarded. An image cannot be part of more than three accepted pairs. An image pair can be accepted if the angle between the two boresight vectors, representing the viewing direction, is between 10° and 30°, and the distance to the target body center varies by no more than 50%. To ensure a sufficient number of pixel correspondences, a limit of 90,000 pixel correspondences is set.

Pixels in large shadowed regions of the target are excluded before obtaining pixel correspondences. This is done by creating a mask through image thresholding and subsequent removal of star effects and smaller shadows using an erode-dilate-erode operation. This step is designed to avoid confusing the learning process with uninformative shadowed regions.

Pixel correspondences are determined by providing the pixel-wise Cartesian coordinates of the second image to a kd-tree and querying it with each Cartesian coordinate of the first image to find the nearest eight neighbors within a limiting distance of $d_{max} = 3\sigma$. The weighted mean of these pixel coordinates determines the corresponding image coordinates in the second image, with weighting based on a Gaussian kernel with variance $\sigma^2$. The lengthscale $\sigma$ is set as the larger of the 90th percentile of the pixel extents of either image. If pixel extents are not provided, they are estimated based on distance and pixel angular size, assuming perpendicular terrain.

Finally, the images are rotated to ensure that the target body's z-axis, typically the rotation axis, is projected upright in each image. Image sizes are increased so that the rotated images fit entirely within the new images, with undetermined border areas filled using the original image's borders. Corresponding pixel coordinates in pairs are also updated to reflect the rotations.

### C. SYNTHETIC IMAGES

In addition to real asteroid and comet images, we generate synthetic images with precise georeferencing to enhance the robustness of the feature extractor against variations in the direction of sunlight. When pairing real images, we observed a similarity in the lighting direction among the created pairs. Notably, the direction of light was not considered during image pair creation, which is left as a potential area for future research due to the challenges in extracting this information from image metadata. To address concerns about lighting robustness, we incorporate synthetic image pairs that exhibit significant variations in lighting. These synthetic images are generated using an existing OpenGL-based image rendering pipeline, as introduced by Knuuttila *et al.* [5], and integrated with SISPO [70].

The renderer leverages camera intrinsics, lighting direction, surface normals at specific locations, and a bidirectional reflectance distribution function (BRDF) to calculate the irradiance ($W/m^2$) received by each pixel. Self-shadowing is handled through shadow mapping. Post-processing includes the addition of background stars, followed by scaling of irradiance values to digital numbers (DNs), assuming arbitrary aperture and optimal integration time. Subsequently, shot noise, moderate readout noise, and dark noise are incorporated. Given the relative brightness of the target object compared to the stars, the stars are mostly imperceptible.

The Hapke 2012 BRDF [79] is employed with default parameter values, which align with values derived from light curve analysis of Bennu [80], as shown in Table 3. However, for each image pair, we introduce randomness by multiplying each parameter value with coefficients drawn from a log-normal distribution with $\sigma^2 = 0.2$. Consistent with the selected target, TAGCAMS [81] is chosen for the camera

Knuuttila *et al.*: CNN-based local features for navigation near an asteroid**TABLE 3.** Default Hapke parameters used in this study

| $\overline{w_0}$ | $B_0$ | $h$ | $g$ | $\overline{\theta}$ |
|---|---|---|---|---|
| 0.31 | 3.9 | 0.11 | -0.32 | 20 |

model. The renderer utilizes a shape model based on the 3.17 m resolution stereo photoclinometry (SPC) derived shape model of Bennu provided by the ORX Altimetry Working Group in 2019 [69]. The original model's vertex count (98,306) led to visible triangles, so we increased the count through smooth interpolation to approximately 2,360,000 vertices. It is worth noting that, subsequent to our data generation stage, various higher resolution shape models derived from laser altimeter data with resolutions of 1.68 m, 0.88 m, and 0.4 m became available as SPICE kernels.

The shape model lacks an accompanying albedo map (texture). In the spirit of data augmentation, we procedurally generate a new albedo map for each image pair. The generated texture, with zero mean, is locally summed to the randomized single scattering albedo $\overline{w_0}$ of the entire body. Our generation scheme is based on Gaussian processes (GPs) [82]. Texture values are generated for each vertex of a low-resolution version of the shape model. We construct the associated covariance matrix using white noise and two different scale Matérn kernels with the shape model's 3D coordinates as inputs. High-resolution shape model vertex texture values are interpolated from the low-resolution values. Additionally, we introduce a high-frequency noise component, modulated by the low-frequency amplitude generated using the same GP covariance matrix. In retrospect, employing three-dimensional Perlin noise [83] or Simplex noise [84] would likely result in a simpler and more efficient texture generation process.

For the first image in each pair, we randomly select the relative orientation, while determining the relative position to ensure that the target fits within the image with some margin. The direction of light is also randomly chosen, ensuring that the phase angle falls within the 0–90° range. For the second image, we perturb the direction of light by introducing two random angles, with their squares uniformly distributed to sample fewer moderate values.

$$\alpha^2 \sim \mathcal{U}(-\alpha_{max}^2, \alpha_{max}^2),$$
$$\beta^2 \sim \mathcal{U}(-\beta_{max}^2, \beta_{max}^2),$$
(17)

where $\alpha$ rotates the direction of light away (or towards) the camera axis, affecting only the phase angle, while $\beta$ rotates it around the camera axis. We limit the maximum rotations to $\alpha_{max} = 45°$ and $\beta_{max} = 180°$. If the resulting phase angle falls outside the 0–90° range, $\alpha$ is resampled.

## V. RESULTS
### A. HYPERPARAMETER OPTIMIZATION
For all network training, we employ the Adam optimizer with a learning rate of 0.001. The batch size for DISK and R2D2-U networks is set to 8, with an image size of $224 \times 224$. The LAFE network, on the other hand, is trained with a batch size of 32 and an image size of $448 \times 448$. In the following sections, we will present the optimization results for DISK, R2D2-U, and LAFE one by one.

#### 1) DISK
Table 4 provides details about the parameters selected for optimization, their initial values, search space, and results. The parameters are categorized into three groups, which correspond to loss function, optimizer, and data augmentation. The only parameter optimized for the Adam optimizer is the weight decay (wd). In terms of loss function, we optimize the false match penalty $\rho_{fp}$, which affects $R_{ij}$ in (12), the sampling cell size $h$, which influences $K_k^u$ in (8), the match distance scale $\theta_M$ in (9), and the pixel error margin $\epsilon$, used as the threshold for true/false matches. The data augmentation group includes parameters such as pixel noise amplitude $\lambda_n$ (section III-C pipeline step 6), synthetic pair maximum rotation $\lambda_r$ and projection $\lambda_p$ in (14), and "synth," which determines whether synthetic images (section IV-C) are used for training or not.

The parameter types include log-uniform (log), uniform (uni), uniform integer (int), or categorical (cat). Different preprocessing methods are applied depending on the parameter type. Log-uniform parameters are transformed into log space before normalization for the GP model. Uniform integer parameters do not require any transformation before normalization, but the values suggested by the surrogate model are rounded before use. To account for nonlinear parameter effects, categorical parameters are label-encoded using integers, avoiding the creation of multiple binary parameters for each label.

The "Initial" column displays the range of values randomly sampled for the first ten trials. These ranges are centered around the values we converged upon after the preliminary testing phase. We designed the full parameter ranges to encompass all reasonable values while avoiding an unnecessarily large search space.

The "Scale" column presents the length scales estimated by the surrogate model during optimization. These length scales provide insight into the parameter's impact on model performance within the specified search space. A longer length scale corresponds to a smaller impact, with a maximum value of 100 indicating negligible impact. The "Result" and "SMMM" columns show the optimization results and the parameter values that maximize the surrogate model mean. Ideally, the optimization result should closely align with the model's optimizing values, especially for parameters with short length scales. Parameter values near the edges of the search space suggest that the search space may have been too narrow.

The M-Score on the validation set for the best DISK model is 34.6, while the maximum mean surrogate model M-Score is 29.1, indicating how much of the performance variation between trials the model attributes to noise. Analyzing the resulting parameter values (see Table 4), it seems that for DISK, further optimization of $\theta_M$ might be possible by exploring

VOLUME 11, 2023                                                                                                                                                           13This work is licensed under a Creative Commons Attribution 4.0 License. For more information, see https://creativecommons.org/licenses/by/4.0/

**TABLE 4.** Optimized Hyperparameter Values for DISK

| Group | Name | Type | Initial | Range | Scale | Result | SMMM |
|---|---|---|---|---|---|---|---|
| loss | $\rho_{fp}$ | uni | 0.23–0.27 | 0.0–0.5 | 24.5 | 0.0154 | 0.00 |
|  | $h$ | cat | 8 | {6, 8, 12} | 1.93 | 8 | 8 |
|  | $\theta_M$ | log | 48–52 | 20–500 | 2.43 | 20.0 | 20.0 |
|  | $\epsilon$ | uni | 1.4–1.6 | 1.0–5.0 | 1.25 | 3.82 | 5.00 |
| opt | wd | log | 0.9e-6–1.1e-6 | 1e-8–1e-3 | 2.82 | 1e-8 | 7.25e-4 |
| data | $\lambda_r$ | uni | 8–12 | 0–20 | 3.74 | 20.0 | 11.3 |
|  | $\lambda_p$ | uni | 0.45–0.55 | 0.2–0.9 | 100.0 | 0.20 | 0.20 |
|  | $\lambda_n$ | uni | 0.08–0.12 | 0.0–0.3 | 8.94 | 0.00 | 0.00 |
|  | synth | cat | false | {false, true} | 3.89 | false | false |

values less than 20. However, when considering the cost of redoing the optimization, this was not deemed worthwhile.

An interesting observation is the discrepancy between the weight decay (wd) and $\lambda_r$ values in the result and SMMM. This implies that rotation data augmentation consumes network capacity, reducing the need for regularization through weight decay. Additionally, the high value for the rotation augmentation may indicate that the scheme, which normalizes scene orientation by rotating the images so that the target object's rotation axis points upward, is not ideal. Parameters $\lambda_p$ and $\rho_{fp}$ appear to have minimal impacts.

Another potential approach to examine the hyperparameter optimization results is to analyze the partial dependence of the optimized metric on different hyperparameter pairs, as estimated by the surrogate model. However, the resulting figures are large and non-essential for presenting our results and are therefore omitted.

### 2) R2D2-U

Table 5 presents the hyperparameter optimization results for R2D2-U. The selected hyperparameters are the same as those for DISK in the optimizer and data augmentation groups, while the parameters related to the loss function are different. We optimize the repeatability weight $\alpha$ and peakiness weight $\beta$ in (3), the acceptable AP threshold $\kappa$ in (6), the cosine similarity window size $n_{rep}$ affecting $P$ in (4), the maximum distance for positive samples $r_{pos}$, and the minimum distance for negative samples $r_{neg}$. Please refer to the discussion related to DISK's results in Table 4 for explanations of common hyperparameter and column meanings.

The optimized R2D2-U M-Score on the validation set is 39.9, outperforming DISK, and its performance appears more stable when re-training it with similar hyperparameter values. Examining the resulting parameter values (see Table 5), it is evident that $\alpha$ and weight decay (wd) are located at the edge of the search space, suggesting potential for further optimization in the future. Similar to DISK, R2D2-U also exhibits a discrepancy between the result and the SMMM for $\lambda_p$ and $\lambda_n$, which seem to impact network capacity and provide regularization, respectively.

We select the optimized R2D2-U model as the authoritative high-performance HAFE model, which serves as a teacher for our lightweight LAFE model. As synthetic Bennu images did not improve performance, LAFE training is also conducted without them. However, we will still use synthetic images to evaluate the performance of the resulting feature extractors when the direction of sunlight changes.

### 3) LAFE

Table 6 provides the hyperparameter optimization results for LAFE. Due to the relative simplicity of the loss function combined with the multitask learning scheme [48], there is no need to optimize any parameters related to the loss function. This allowed us to use the negated validation loss as the optimization metric. Regarding data augmentation, none of the previous parameters are required, as single images are used for training. We have included two parameters: student image random gain $\lambda_g^{st}$ and noise SD $\lambda_\sigma^{st}$ (introduced at the end of section III-C). There is also a new group for network model-related parameters, "arch," determining which backbone to use (mn2 for MobileNetV2, mn3 for MobileNetV3, and en0 for EfficientNet-B0), and "desc-se," determining if the descriptor head should use squeeze-excitation or not.

The negated validation loss achieved by the best model was 1.014, while the maximum mean value given by the surrogate model was 0.803. Examining the length scales of $\lambda_g^{st}$ and $\lambda_\sigma^{st}$, it appears that adding noise to images used as input by the student network during training has a negligible effect on the resulting network. Interestingly, the resulting model from the optimization is based on MobileNetV3, while the surrogate model suggests that MobileNetV2 might be a better choice. The length scale of the backbone selection is very small at 0.412, prompting us to train the predicted best model, resulting in a negated validation loss of 0.877. Although this metric is lower than that of the best model, we will include this surrogate model-suggested feature extractor in our subsequent analysis, referring to it as LAFE-SM.

### B. EVALUATION

To assess the expected performance of the resulting feature extractors in our specific use case of visual navigation in close proximity to asteroids, we will evaluate them by calculating typical feature matching metrics, such as M-Score, MMA, and pixel localization error. Additionally, we will estimate the relative poses between the asteroid and spacecraft based on the matched features, assuming knowledge of their 3D coordinates. This involves initial geometric verification of matches using RANSAC and then refining the pose through a simplified bundle adjustment (BA) scheme, where adjustments are made solely to the pose parameters. To enhance robustness against high reprojection errors, we employ a pseudo-Huber loss function. The local 3D coordinates are derived from depth maps extracted for Eros, 67P/C–G OSINAC, Itokawa, and synthetic image datasets. Ground truth relative poses are obtained using the same pose estimation pipeline, but instead of feature matches, it relies on ground truth pixel correspondences. For feature matches, the maximum allowable reprojection error for RANSAC is set to 5 pixels, while for ground truth poses, it is reduced to 0.75 pixels. When the number of pixel correspondences exceeds 20,000, we subsample by dropping every $k = floor(n/10000)$ pixel correspondences, ensuring a minimum of 10,000 correspondences for pose estimation.

**TABLE 5.** Optimized Hyperparameter values for R2D2-U

| Group | Name | Type | Initial | Range | Scale | Result | SMMM |
|---|---|---|---|---|---|---|---|
| loss | $\alpha$ | uni | 0.23–0.27 | 0.1–1.0 | 3.40 | 0.10 | 0.10 |
|  | $\beta$ | uni | 0.18–0.22 | 0.05–0.5 | 24.1 | 0.50 | 0.50 |
|  | $\kappa$ | uni | 0.58–0.62 | 0.5–0.99 | 0.903 | 0.596 | 0.587 |
|  | $n_{rep}$ | int | 23–25 | 16–32 | 100.0 | 21 | 32 |
|  | $r_{pos}$ | int | 1–2 | 1–5 | 2.17 | 1 | 1 |
|  | $r_{neg}$ | int | 9–11 | 6–20 | 2.58 | 19 | 14 |
| opt | wd | log | 0.9e-6–1.1e-6 | 1e-8–1e-3 | 0.728 | 1e-8 | 1e-8 |
| data | $\lambda_r$ | uni | 8–12 | 0–20 | 46.3 | 17.4 | 20.0 |
|  | $\lambda_p$ | uni | 0.45–0.55 | 0.2–0.9 | 3.17 | 0.701 | 0.259 |
|  | $\lambda_n$ | uni | 0.08–0.12 | 0.0–0.3 | 2.02 | 0.00 | 0.0790 |
|  | synth | cat | false | {false, true} | 2.34 | false | false |

**TABLE 6.** LAFE Optimized Hyperparameters

| Group | Name | Type | Initial | Range | Scale | Result | SMMM |
|---|---|---|---|---|---|---|---|
| model | arch | cat | mn2 | {mn2, mn3, en0} | 0.412 | mn3 | mn2 |
|  | desc-se | cat | true | {true, false} | 3.40 | false | false |
| opt | wd | log | 0.9e-8–1.1e-8 | 1e-9–1e-5 | 1.45 | 1.78e-9 | 1.71e-8 |
| data | $\lambda_g^{st}$ | uni | 1.08–1.12 | 1.0–1.3 | 100.0 | 1.29 | 1.30 |
|  | $\lambda_\sigma^{st}$ | uni | 0.01–0.02 | 0.0–0.1 | 100.0 | 0.0067 | 0.00 |

Bennu and 67P/C–G NAVCAM datasets are excluded from the evaluation since real image pairs were not available for them. The Itokawa dataset was initially reserved solely for the final evaluation of the feature extractors and was not used for hyperparameter optimization. Even though we considered including the synthetic image dataset in the training process, the final feature extractors were not trained with it. The inclusion of synthetic images in hyperparameter optimization is unlikely to introduce subtle overfitting, as the dataset was excluded from the training set. All feature extractors were trained using the Eros and 67P/C–G OSINAC datasets, which may lead to some bias toward these datasets.

In addition to our DISK and R2D2-U HAFE models (HAFE-DISK, HAFE-R2D2), and the two LAFE models (regular LAFE and LAFE-SM), we also include RootSIFT in our evaluation. RootSIFT consistently outperformed regular SIFT and AKAZE, which performed at a similar level to regular SIFT. RootSIFT is essentially SIFT with the descriptor elements subjected to the square root operation and the resulting vectors normalized to unit length [85]. To ensure comparability in feature performance, we extract the same number of features with all the feature detectors. We set the number of layers per octave (halving of image size) to 4 for both RootSIFT and our learning-based extractors.

Due to challenges encountered in training the original R2D2 feature extractor with our dataset, we cannot directly compare our proposed feature extractors with the original R2D2. Nevertheless, we include the original R2D2 feature extractor, trained on urban scenery (R2D2-Orig, model file r2d2_WASF_N16.pt), in our evaluation as a reference baseline to emphasize the significance of training the network with asteroid data. Additionally, we developed our own version of R2D2 (R2D2-VGG), closely resembling the original but using our dataset. It's worth noting that there are some differences between our implementation and the original, including variations in data augmentation and loss function annealing. Consequently, we cannot conclusively assert that our HAFE-R2D2 model outperforms the original R2D2.

The results are reported separately for each dataset, as shown in Table 7, Table 8, Table 9, and Table 10. The image pairs within each dataset are categorized as "easy" or "hard" based on the magnitude of the change in viewing angle $|\varphi|$. The synthetic dataset, however, is classified based on both the magnitude of the change in phase angle $|\alpha|$ and the light direction $|\beta|$, as defined in (17). An image pair is considered "easy" if $|\varphi| < 15°$, and "hard" otherwise. For the synthetic dataset, both $|\alpha| < 20°$ and $|\beta| < 30°$ must hold for the pair to be deemed "easy."

For clarity and conciseness, we include only M-Score, orientation estimation failure rate, and orientation error 50- and 85-percentiles as metrics in these tables. Orientation estimation is considered to have failed for an image pair if the result comprises fewer than 12 features with a reprojection error of less than five pixels or if the orientation error exceeds 20°. To enable comparability of orientation error percentiles across different failure rates, we treat failures as having an arbitrarily large orientation error. The best-performing method for each

**TABLE 7.** Results on Eros dataset

| Method | Subset | M-Score | Fail % | Orient. Err p50 | p85 |
|---|---|---|---|---|---|
| HAFE-DISK | easy | 47.5 | 9.67 | 0.307 | 1.27 |
| HAFE-R2D2 | easy | 54.5 | 5.20 | 0.205 | 0.682 |
| LAFE | easy | 50.0 | 6.26 | 0.223 | 0.737 |
| LAFE-ME | easy | 49.7 | 7.55 | 0.221 | 0.781 |
| R2D2-Orig | easy | 37.8 | 16.3 | 0.227 | |
| R2D2-VGG | easy | 48.5 | 8.94 | 0.211 | 0.755 |
| RootSIFT | easy | 15.6 | 22.8 | 0.291 | |
| HAFE-DISK | hard | 33.9 | 21.5 | 0.466 | |
| HAFE-R2D2 | hard | 39.9 | 11.8 | 0.316 | 3.55 |
| LAFE | hard | 34.6 | 15.0 | 0.369 | 16.2 |
| LAFE-ME | hard | 33.0 | 17.6 | 0.357 | |
| R2D2-Orig | hard | 20.2 | 40.6 | 0.551 | |
| R2D2-VGG | hard | 34.3 | 19.5 | 0.319 | |
| RootSIFT | hard | 9.74 | 40.2 | 0.634 | |

$n_{easy} = 1875, n_{hard} = 3961$

**TABLE 8.** Results on 67P/C–G dataset

| Method | Subset | M-Score | Fail % | Orient. Err p50 | p85 |
|---|---|---|---|---|---|
| HAFE-DISK | easy | 26.1 | 25.6 | 0.707 | |
| HAFE-R2D2 | easy | 29.6 | 16.0 | 0.545 | |
| LAFE | easy | 27.6 | 16.4 | 0.555 | |
| LAFE-ME | easy | 26.6 | 20.4 | 0.607 | |
| R2D2-Orig | easy | 19.9 | 32.6 | 0.661 | |
| R2D2-VGG | easy | 25.2 | 22.6 | 0.544 | |
| RootSIFT | easy | 8.77 | 46.5 | 1.50 | |
| HAFE-DISK | hard | 14.8 | 36.7 | 1.75 | |
| HAFE-R2D2 | hard | 18.9 | 22.4 | 0.935 | |
| LAFE | hard | 16.6 | 24.4 | 1.09 | |
| LAFE-ME | hard | 15.2 | 30.0 | 1.16 | |
| R2D2-Orig | hard | 8.82 | 53.3 | | |
| R2D2-VGG | hard | 15.3 | 31.9 | 1.09 | |
| RootSIFT | hard | 3.91 | 65.6 | | |

$n_{easy} = 1371, n_{hard} = 5362$

metric and dataset is highlighted in gray.

For the Eros, 67P/C–G, and synthetic datasets, HAFE-R2D2 dominates in terms of M-Score, failure rate, and orientation error percentiles. The synthetic dataset produces unreliable results for the "easy" subset due to the limited number of samples therein. This is a consequence of the sampling

**TABLE 9.** Results on Itokawa dataset

| Method | Subset | M-Score | Fail % | Orient. Err p50 | p85 |
|---|---|---|---|---|---|
| HAFE-DISK | easy | 41.9 | 0.355 | 0.555 | 1.95 |
| HAFE-R2D2 | easy | 42.1 | 2.13 | 0.400 | 1.69 |
| LAFE | easy | 44.0 | 0.355 | 0.386 | 1.16 |
| LAFE-ME | easy | 47.5 | 0.355 | 0.350 | 1.00 |
| R2D2-Orig | easy | 40.0 | 0.355 | 0.357 | 0.804 |
| R2D2-VGG | easy | 50.0 | 0.000 | 0.319 | 0.778 |
| RootSIFT | easy | 10.9 | 13.1 | 0.380 | 3.08 |
| HAFE-DISK | hard | 28.6 | 4.31 | 0.887 | 3.30 |
| HAFE-R2D2 | hard | 31.5 | 4.09 | 0.623 | 2.84 |
| LAFE | hard | 31.3 | 0.681 | 0.603 | 1.97 |
| LAFE-ME | hard | 32.5 | 0.908 | 0.553 | 1.57 |
| R2D2-Orig | hard | 22.3 | 3.41 | 0.649 | 1.69 |
| R2D2-VGG | hard | 37.6 | 0.795 | 0.485 | 1.22 |
| RootSIFT | hard | 4.67 | 45.6 | 2.68 | |

$n_{easy} = 367, n_{hard} = 1129$

**TABLE 10.** Results on Synthetic dataset

| Method | Subset | M-Score | Fail % | Orient. Err p50 | p85 |
|---|---|---|---|---|---|
| HAFE-DISK | easy | 46.9 | 0 | 0.411 | 0.826 |
| HAFE-R2D2 | easy | 61.5 | 0 | 0.158 | 0.269 |
| LAFE | easy | 61.1 | 0 | 0.120 | 0.207 |
| LAFE-ME | easy | 61.6 | 0 | 0.148 | 0.236 |
| R2D2-Orig | easy | 53.3 | 0 | 0.120 | 0.234 |
| R2D2-VGG | easy | 52.5 | 0 | 0.132 | 0.201 |
| RootSIFT | easy | 22.2 | 0 | 0.170 | 0.283 |
| HAFE-DISK | hard | 3.49 | 77.4 | | |
| HAFE-R2D2 | hard | 6.05 | 22.0 | 2.49 | |
| LAFE | hard | 5.75 | 23.3 | 2.75 | |
| LAFE-ME | hard | 5.07 | 36.4 | 4.05 | |
| R2D2-Orig | hard | 3.56 | 75.2 | | |
| R2D2-VGG | hard | 4.10 | 66.9 | | |
| RootSIFT | hard | 0.800 | 93.0 | | |

$n_{easy} = 47, n_{hard} = 7916$

distribution that prioritizes significant changes in lighting, resulting in numerous "hard" samples. Consequently, the "hard" subset exhibits a bias toward more challenging image pairs, leading to excessively pessimistic performance metric values. On these datasets, LAFE performs slightly worse than HAFE-R2D2 but outperforms all other feature extractors, including those with more sophisticated architectures such as HAFE-DISK and R2D2-VGG. Interestingly, R2D2-VGG outperforms the others on the Itokawa dataset, while both LAFE and LAFE-ME significantly outperform HAFE-R2D2, the model used to train both extractors. The small sample size may contribute to the observed outcome, although further investigation is required to determine the exact cause with certainty.

As expected, the original R2D2, trained solely on urban scenery, performs poorly compared to other learning-based feature extractors trained on relevant data. Nevertheless, it still outperforms RootSIFT.

The primary objective of this work has been to develop a lightweight feature extractor for navigation near asteroids. Therefore, Figures 5–8 focus solely on LAFE performance metrics. The first three figures illustrate how the distribution (including the median) of the metrics is influenced by $|\varphi|$. The samples are grouped into bins based on $|\varphi|$, and intra-bin empirical distributions are estimated using Gaussian kernels, presented as violin plots. For the synthetic dataset (Fig. 8), we display the medians in specific bins determined by $|\alpha|$ and $|\beta|$. The metrics include M-Score, MMA, LE, and orientation error. For orientation error median calculations, estimation failures are considered as arbitrarily large errors. Due to the challenge in interpreting the mAP metric, it is excluded from the figures to enhance readability.

The violin plots reveal substantial variation in performance among different image pairs within the view angle change bins. This variability could stem from varying lighting conditions, which we were unable to extract from these images during this study. Notably, this variance is more pronounced in the 67P/C–G dataset, while it is relatively subdued in the Itokawa dataset. Based on the median orientation error in the synthetic dataset, it appears that LAFE performs reasonably well up to a 50° change in phase angle ($\alpha$) and a 50° change in the direction of light ($\beta$). However, this should be verified with real asteroid imagery and properly extracted lighting-direction information.

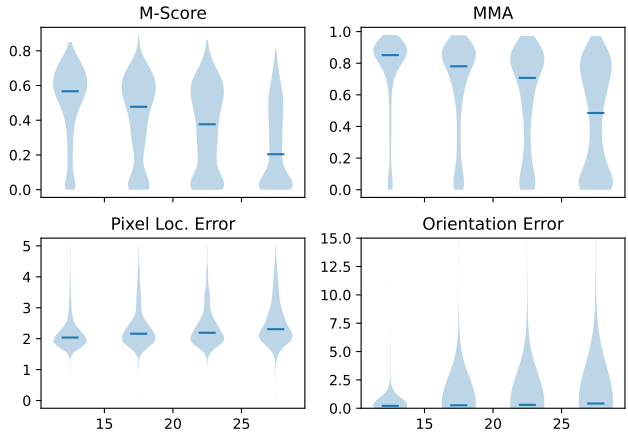

**FIGURE 5.** Eros: LAFE performance metrics on the vertical axes versus changes in viewing angle in degrees.

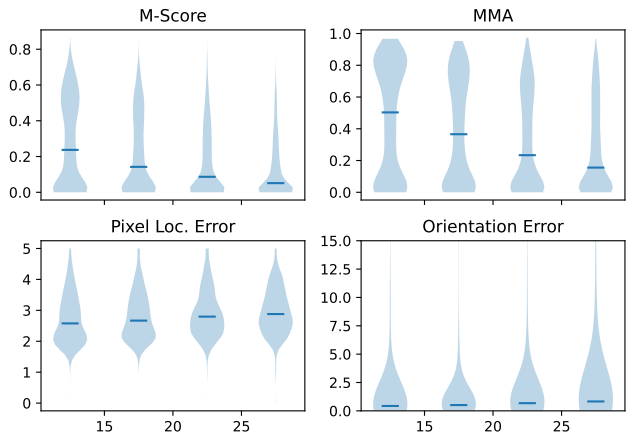

**FIGURE 6.** 67P/C–G: LAFE performance metrics on the vertical axes versus changes in viewing angle in degrees.

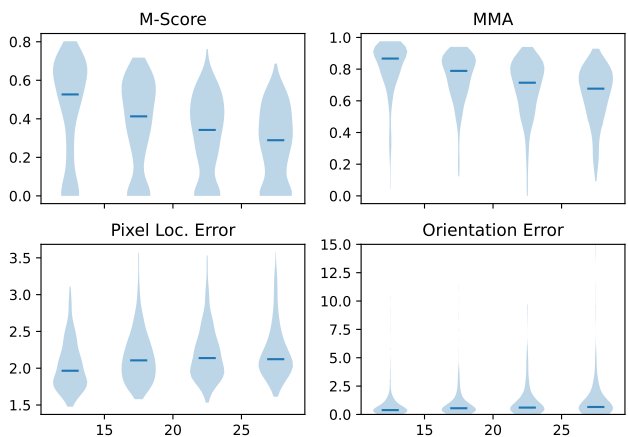

**FIGURE 7.** Itokawa: LAFE performance metrics on the vertical axes versus changes in viewing angle in degrees.

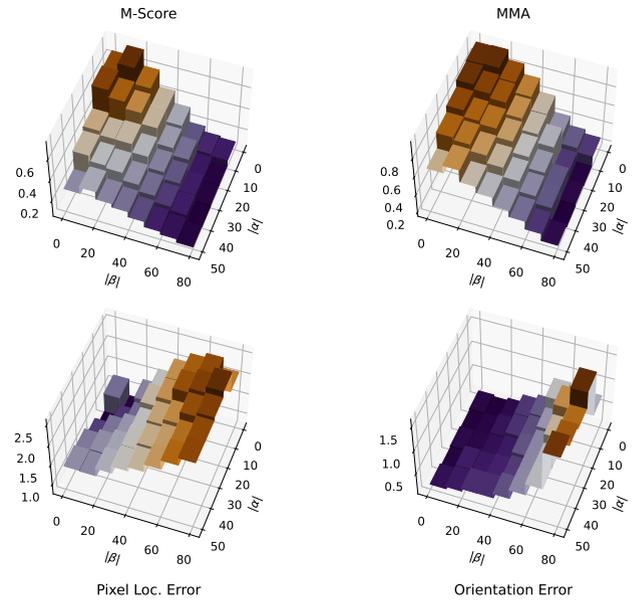

**FIGURE 8.** Synthetic data: LAFE median performance versus changes in lighting in degrees. See text for details.

Fig. 9 provides an example of an image pair from the 67P/C–G dataset. It displays a corresponding LAFE detection map along with feature matches that have undergone geometric validation. Observations suggest that the detections primarily focus on smaller shadows, which may raise some level of concern since shadows tend to shift with changes in lighting conditions. However, under normal circumstances, small shadows typically move only short distances, resulting in reprojection errors of 5 pixels or less.

To assess the computational performance of LAFE, its PyTorch model was exported to the ONNX format [86], which was then utilized in conjunction with ONNX Runtime [87] to extract features at a single scale from 756 Itokawa images. Memory usage and execution time (excluding initialization) were measured on an Ultra96 board [88] equipped with a Xilinx Zynq UltraScale+ MPSoC, model ZU3EG. The primary four-core ARM Cortex A53 CPU, operating at 1200 MHz, was utilized for processing. Throughout the experiment, memory usage remained below 400 MB, and with all four CPU cores in use, the average execution time per image was 0.66 seconds. If only one core was used, the average time was 1.34 seconds. For comparison, we also measured the single-core performance of the HAFE-R2D2 model. In this case, memory usage was approximately 1200 MB, and the average execution time for the first 12 images was 54.1 seconds. The breakdown of tasks performed for each image, along with the respective time allocations, were as follows: loading the image from the filesystem (6.4%), resizing it to a resolution of 512x512 (2.3%), performing LAFE inference (84.1%), extracting sparse features (6.1%), and saving the features back to the filesystem (0.4%). The remaining 0.7% accounted for managing the image processing loop, ensuring a cumulative total of 100%.

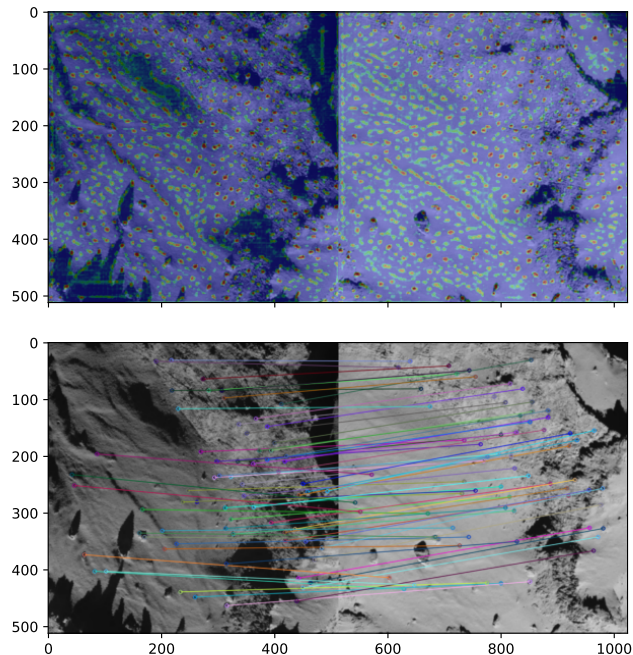

**FIGURE 9.** An example 67P/C–G OSINAC image pair with overlaid LAFE feature detection map (top) and successfully matched features (bottom). Image credit: ESA

**TABLE 11.** AV-ASLFeat and HAFE M-Score comparison

| Study | Body | R2D2-Orig | Proposed | Increase |
|---|---|---|---|---|
| AstroVision | Itokawa | 9.4 | 11.0 | 17.0% |
| ours | Itokawa - easy | 40.0 | 42.1 | 5.25% |
| ours | Itokawa - hard | 22.3 | 31.5 | 41.3% |
| AstroVision | 67P/C–G | 16.5 | 23.9 | 44.8% |
| ours | 67P/C–G - easy | 19.9 | 29.6 | 48.7% |
| ours | 67P/C–G - hard | 8.82 | 18.9 | 114% |

We did not conduct experiments with multiscale extraction. However, we can provide a rough estimation by assuming a constant per-pixel computation time. If we were to extract features ranging from images of size 512x512 to 128x128, with four scale levels per octave, the total pixel count would be equivalent to an image of size 925x925. Based on the pixel count ratio, the estimated time using a single core would be 4.37 seconds.

## VI. CONCLUSION

In this study, we have successfully developed a lightweight asteroid feature extractor (LAFE) designed for onboard execution on a reasonably capable CPU, such as the Xilinx Zynq 7000-series SoC or the more capable Xilinx Zynq Ultrascale+ MPSoC. Real-time performance is achievable for mission profiles with slow dynamics that require infrequent feature extraction as permitted by CPU execution. For higher-frequency feature extraction, hardware acceleration can be employed by utilizing Xilinx Vitis AI [89] to convert and fine-tune the model for execution on an 8-bit DPU (Deep Learning Processing Unit) provided by Xilinx, which can be implemented on the FPGA section of an Ultrascale+ MPSoC.

During our research, we also trained a high-performance asteroid feature extractor (HAFE), which served as the teacher for LAFE. HAFE incorporates several improvements over the state-of-the-art R2D2 feature extractor, including the use of a U-Net backbone, a warmup period for the loss function parameter $\kappa$ in (6), optimized loss function weights $\alpha$ and $\beta$ in (2), and a single-channel output for the reliability head instead of using softmax and two channels. Through hyperparameter optimization, we were able to reliably compare two state-of-the-art feature extractors and determined that our incrementally improved R2D2-U extractor outperformed DISK on asteroid imagery. Additionally, we compared three lightweight architectures for LAFE and found that MobileNetV3 outperformed both EfficientNet-B0 and MobileNetV2.

Concurrent to our research, Driver *et al.* [40] introduced an intriguing ASLFeat-based feature extractor, AV-ASLFeat (originally ASLFeat-CVGBEDTRPJMU), along with a valuable small body dataset called AstroVision. However, a notable limitation of their work is the inability to compare AV-ASLFeat with other learning-based feature extractors trained on relevant data. Their comparisons were limited to traditional feature extractors and learning-based extractors trained on various terrestrial datasets. Unfortunately, their dataset and trained model were not available at the time of writing, preventing direct comparisons between our proposed feature extractors, HAFE and LAFE, with AV-ASLFeat.

While Driver *et al.* utilized a true match limit of 5 pixels and employed metrics such as recall, precision, and orientation error, which align with our M-Score, MMA, and orientation error, respectively, the differences in image pairing, pose estimation algorithm, and allowed feature count per image make direct comparison of results impossible.

However, since Driver *et al.* also included the original R2D2 model in their results, it may be possible to gain some insight into the relative performance of AV-ASLFeat and HAFE by comparing their M-Score/Recall to the original R2D2 evaluated in their respective studies. Table 11 provides a summary of the relative improvements of all proposed methods over R2D2 for different subsets of data. It is important to note that even this form of comparison, as highlighted by the Itokawa data and its easy and hard subsets, is severely limited due to the significant influence of image pairing on the relative gains in M-Score. Future work should include consolidating and directly comparing our proposed methods with AV-ASLFeat.

Regarding future work, several avenues remain unexplored for potentially enhancing our feature extraction performance. Notably, preprocessing the datasets to include information about the direction of light could improve image pairing, resulting in better training data and more informative evaluation results, which can be presented as performance metrics as a function of the magnitude of change in the direction of light. Additionally, investigating the impact of descriptor dimensionality on performance and computational resource usage is important, as the choice of 128 dimensions was based solely on convention. Other unoptimized parameters, such as learning rate (currently set at 1e-3), match filtering using the ratio test, and non-maximum suppression radius (currently set at 3 pixels), warrant exploration. It may also be instructive to test the feature extractors by replacing the feature detection component with classical methods, such as Harris corners, for instance.

Eliminating the need to rotate images based on an asteroid rotation model prior to feature extraction could potentially be achieved using a spatial transformer [52].

Another promising avenue for future work that could result in full illumination invariance could be to investigate a two-stage feature extraction architecture consisting of a depth-estimating first stage and a feature-extracting second stage, connected by a 3D spatial transformer. The first stage could be directly trained with depth information and could include outputs for depth uncertainty and possibly an albedo estimate.

Furthermore, in addition to improving local feature extraction, the shape-model-based SPL algorithm [5] used for absolute navigation could be significantly improved by substituting the AKAZE features with a lightweight feature extractor specifically trained for this purpose.

A crucial next step involves enhancing LAFE by incorporating a global descriptor head and integrating the resulting network with a SLAM algorithm, allowing the evaluation of navigation performance on a specified mission profile using appropriate simulation software.

## ACKNOWLEDGMENT

The authors wish to acknowledge Aalto Science-IT project and CSC - IT Center for Science, Finland, for computational resources.

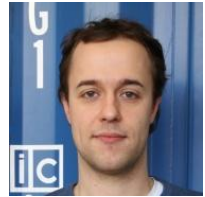

**ANTTI KESTILÄ** is a researcher at the Finnish Meteorological Institute (FMI). He has an M.Sc. from Aerospace Engineering from TU Delft and a Ph.D. in Space Technology from Aalto University. Dr. Kestilä has been working on several small satellite missions such as Aalto University's Aalto-1 and Suomi 100 CubeSats, the AIDA/AIM Asteroid Spectral Imaging Mission ASPECT CubeSat, and later the HERA mission APEX CubeSat. His research interests include small satellite missions and space instrumentation, as well as novel space materials.

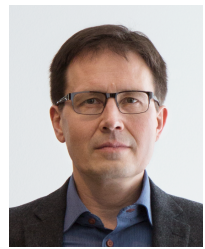

**ESA KALLIO** is an Associate Professor of Space Science and Technology at Aalto University, Helsinki, Finland. He received his Ph.D. (1996) in space physics from the University of Helsinki, Finland, with a focus on ASPERA/Phobos-2 ion spectrometer and its measurements at Mars. His main area of research has involved space plasma instruments and interpretation of their observations with computer simulations. He has been involved as a Coinvestigator in several space instrument projects: ASPERA-3/Mars Express, ASPERA-4/Venus Express, ICA/Rosetta, and PEP/JUICE. His recent research interests include CubeSats and their space instrumentation. Most recently, he led the development of the Suomi 100 CubeSat.

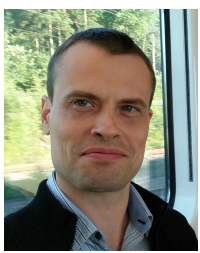

**OLLI KNUUTTILA** is a doctoral researcher of Space Technology at Aalto University. He received his M.Sc. (2008) in Computer and Information Science from Helsinki University of Technology. After a career in online retail technology, his interests led to a full commitment to advance artificial intelligence, and particularly optical navigation in future small satellites for deep space exploration. He was responsible for optical navigation algorithms for the now defunct APEX project, where a daughter spacecraft was to be released from the Hera mothership after arrival at the binary asteroid 65803 Didymos.